\documentclass{article}

\PassOptionsToPackage{numbers, compress}{natbib}

\usepackage[preprint]{neurips_2026}

\usepackage[utf8]{inputenc}
\usepackage[T1]{fontenc} 
\usepackage{hyperref}   
\usepackage{url}         
\usepackage{booktabs}       
\usepackage{amsfonts}       
\usepackage{nicefrac}      
\usepackage{microtype} 
\usepackage{xcolor} 
\definecolor{igpositive}{RGB}{0,128,0}
\definecolor{goldanswer}{RGB}{0,80,180}
\usepackage{amsmath} 
\usepackage{algorithm} 
\usepackage{algorithmic} 
\usepackage{colortbl} 
\usepackage{multirow} 
\usepackage{graphicx,lipsum,afterpage,subcaption}
\usepackage{makecell}

\title{SD-Search: On-Policy Hindsight Self-Distillation for Search-Augmented Reasoning}

\author{%
  Yufei Ma\thanks{Equal Contribution}, \quad 
  Zihan Liang\footnotemark[1], \quad
  Ben Chen\footnotemark[1] \thanks{Corresponding author}, \quad Zhipeng Qian, \quad Huangyu Dai \\
  \textbf{Lingtao Mao}, \quad \textbf{Xuxin Zhang}, \quad \textbf{Chenyi Lei}, \quad \textbf{Wenwu Ou} \\
  \\
  Kuaishou Technology \\
  \texttt{benchen4395@gmail.com}
}

\begin{document}

\maketitle

\begin{abstract}
Search-augmented reasoning agents interleave internal reasoning with calls to an external retriever, and their performance relies on the quality of each issued query. 
However, under outcome-reward reinforcement learning, every search decision in a rollout shares the same trajectory-level reward, leaving individual queries without step-specific credit.
Recent process-supervision approaches address this gap by drawing step-level signals from outside the policy, relying either on a much larger teacher model, or on sub-question annotations produced by a stronger external system.
In contrast, we propose \textbf{SD-Search}, which derives step-level supervision \emph{from the policy itself} through on-policy hindsight self-distillation, requiring neither an external teacher nor additional annotations.
In SD-Search, a single model plays two roles that differ only in conditioning: a student that sees only the context available at inference time, and a teacher that additionally conditions on a compact hindsight block summarizing the search queries and final outcomes of a group of rollouts sampled from the same question.
Since the teacher knows how each rollout unfolded and which ones succeeded, its query distribution implicitly marks which decisions were worth making, and the student is trained to recover this behavior by minimizing the token-level Jensen--Shannon divergence to the teacher at search-query positions.
This layers a dense, step-level signal on top of GRPO's coarse trajectory reward. 
Crucially, this signal is produced by the policy itself within the standard RL training loop, without external model inference, auxiliary annotation pipeline, or additional training stage.
On seven single-hop and multi-hop QA benchmarks, SD-Search reaches $0.428$ average Exact Match with Qwen2.5-3B, matching the leading process-supervision baseline without its external teacher, and $0.476$ with Qwen2.5-7B, surpassing all outcome-reward and process-supervision baselines on the average.
\end{abstract}

\section{Introduction}
\label{sec:intro}

\par
Large language models (LLMs) have achieved strong reasoning performance, yet on knowledge-intensive tasks they remain limited by what was available at pretraining and cannot incorporate information that emerged afterwards~\citep{lewis2020naiverag, openai2023gpt4,qwen2025,borgeaud2022retro,jason2022cot}.
A recent line of work addresses this by training LLMs as search-augmented reasoning agents that interleave internal reasoning with calls to an external retriever, autonomously deciding when to search and what to query for~\citep{jin2025searchr1, chen2025research, shi2025autorefine,timo2023toolformer,yao2022react}.
These agents are typically optimized with reinforcement learning such as GRPO~\citep{shao2024deepseekmath,schulman2017ppo}, using a trajectory-level reward derived from the correctness of the final answer.
This formulation has driven steady progress on standard QA benchmarks, but individual search decisions, which largely determine what evidence the model gets to reason over, receive no step-specific credit.

\par
Several recent methods attempt to close this gap by injecting step-level supervision into the training signal.
Thinker~\citep{xu2025thinker} decomposes problems into sub-questions with a 72B teacher, then supervises a student through SFT on the resulting decomposed trajectories, while StepSearch~\citep{wang2025stepsearch} derives step-wise rewards from decomposition annotations produced by GPT-4o~\citep{openai2023gpt4}.
These approaches deliver clear improvements, but they share a common dependency: the step-level signal is imported from a stronger external system, through either distillation data or auxiliary annotations, which adds cost and restricts applicability to settings where such resources exist.

\begin{figure}[!htbp]
    \centering
    \includegraphics[width=0.85\linewidth]{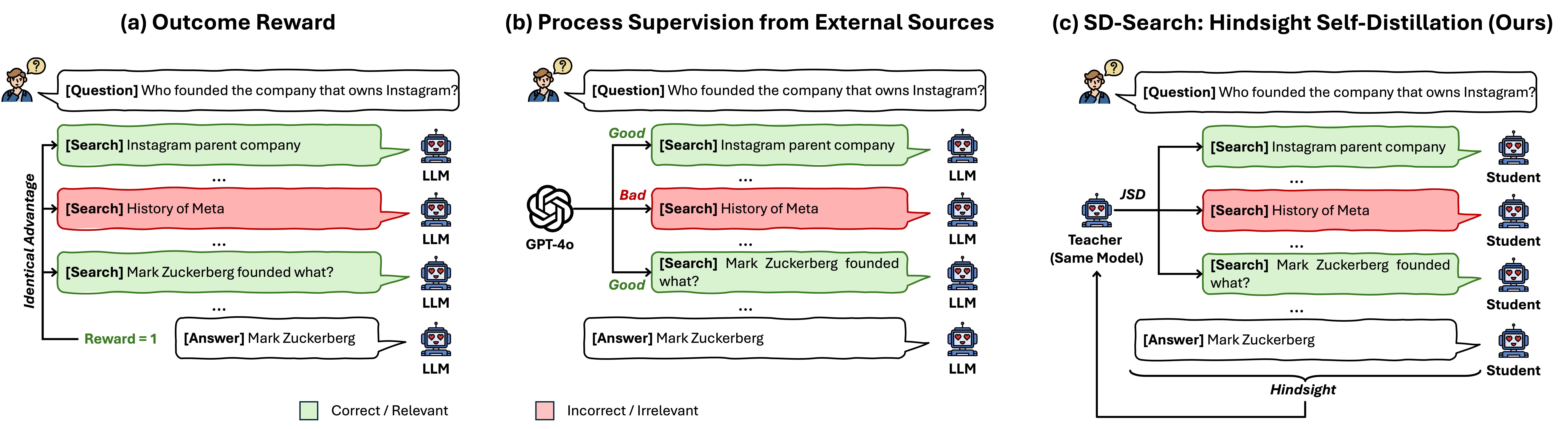}
    \caption{Three paradigms for supervising search decisions. \textbf{(a) Outcome Reward (e.g., AutoRefine):} a single trajectory-level reward, failing to distinguish between good and bad individual queries. \textbf{(b) Process Supervision (e.g., Thinker, StepSearch):} step-level signals imported from external sources (a 72B teacher or GPT-4o-generated sub-question annotations). \textbf{(c) SD-Search (Ours):} on-policy hindsight self-distillation produces step-level supervision directly from the policy itself, without external resources.}
    \label{fig:motivation}
\end{figure}

\par
We make a different observation. 
The policy itself, once given access to how its rollouts actually unfolded and which ones succeeded, is in a much better position to judge which earlier decisions were worth making than the same policy at decision time.
This suggests that the step-level signal from external sources can instead be recovered from the policy's own predictions under an appropriately constructed hindsight context.
Building on this, we propose \textbf{SD-Search} (Figure~\ref{fig:motivation}), a reinforcement learning framework that supplies step-level supervision through \emph{on-policy hindsight self-distillation}.
The signal is derived entirely from the policy itself, not requiring external teacher or annotations beyond the question-answer pairs used for outcome-reward training.
The policy is queried under two conditioning contexts on the same trajectory: a \emph{student} context, which contains only the prefix available at inference time, and a \emph{teacher} context, which additionally exposes a compact hindsight summary of the search queries and outcome labels of a group of rollouts sampled from the same question.
With hindsight, the teacher's predictive distribution over the search-query tokens implicitly encodes which queries were worth issuing.
The student is then aligned to this distribution via a token-level Jensen-Shannon divergence at the query positions, providing a dense step-level signal that complements the coarse trajectory-level advantages of GRPO.
This group-level conditioning is a deliberate design choice. 
Instead of only showing teacher the single rollout being supervised, we expose it to a contrast between successful and unsuccessful rollouts. 
This comparison sharpens the per-token target and ensures the hindsight signal remains informative, even if the supervised rollout ends in failure.

\par
We evaluate SD-Search on seven single-hop and multi-hop QA benchmarks with both Qwen2.5-3B and Qwen2.5-7B~\citep{qwen2025}.
At the 3B scale, SD-Search improves over the strongest outcome-reward baselines, AutoRefine~\citep{shi2025autorefine} and MR-Search~\citep{xiao2025mrsearch}, by $2.3$ and $1.4$ points in average Exact Match respectively, and matches the leading process-supervision baseline Thinker~\citep{xu2025thinker} without its 72B teacher or supervised distillation stage.
At the 7B scale, SD-Search surpasses every baseline we compare against, exceeding AutoRefine by 2.1 and Thinker by 2.4 points on the seven-benchmark average.
Our contributions are summarized as follows:
\begin{itemize}
    \item We propose \textbf{SD-Search}, a framework that supplies step-level supervision for search decisions through on-policy hindsight self-distillation, using no external teacher and no annotations beyond those required for outcome-reward training.
    \item We introduce a token-level Jensen-Shannon objective that aligns the policy's query distribution at each search step with the distribution produced under hindsight conditioning. The objective integrates into GRPO by adding a single auxiliary forward pass and one loss term, leaving the advantage estimator untouched.
    \item We conduct extensive experiments on seven QA benchmarks at both 3B and 7B scales, showing consistent gains over outcome-reward baselines and matching or surpassing process-supervision baselines, with ablations that isolate the contribution of each design.
\end{itemize}

\section{Method}
\label{sec:method}
\par
\begin{figure*}[!htbp]
    \centering
    \includegraphics[width=0.85\linewidth]{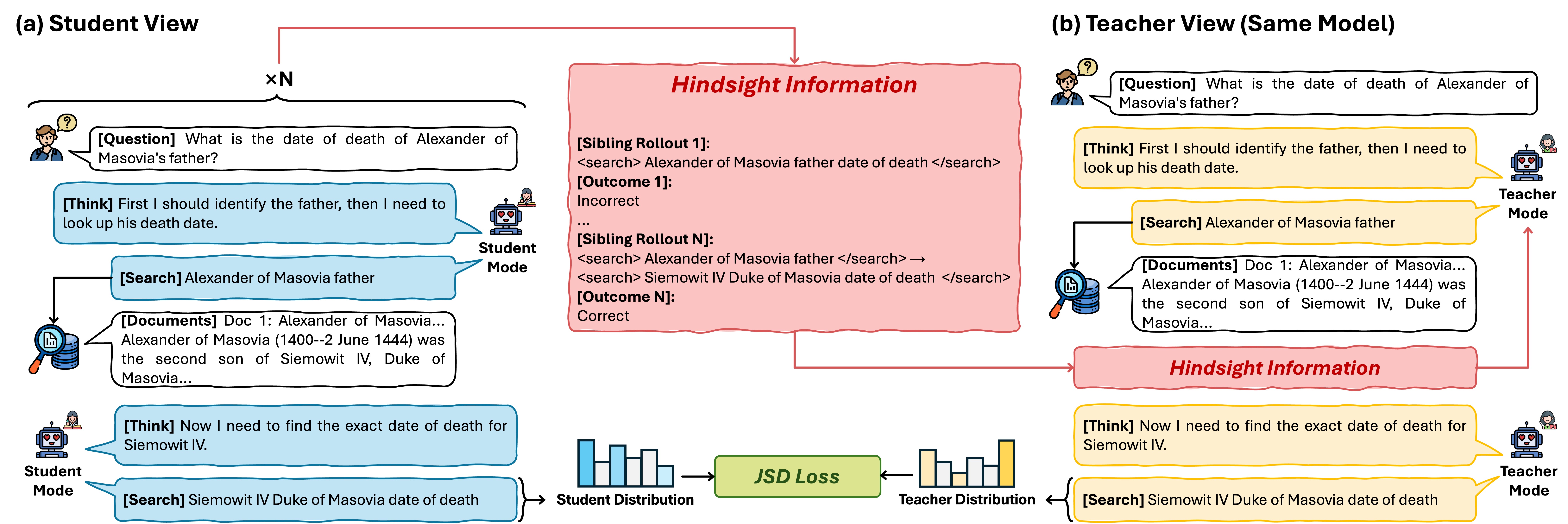}
    \caption{Method overview of SD-Search. The policy acts as its own teacher by conditioning on a \textbf{Hindsight Information} block containing the search trajectories and outcomes of sibling rollouts. The student (a), restricted to standard inference-time context, is aligned to the teacher (b) distribution at search-query positions using a token-level JSD loss, providing dense step-level supervision.}
    \label{fig:method}
\end{figure*}

This section describes the components of \textbf{SD-Search}.
We first review the trajectory format and the GRPO objective inherited from prior work (\S\ref{sec:prelim}).
Then we describe how an asymmetric student-teacher pair is constructed from the same policy by varying only the conditioning context (\S\ref{sec:hindsight}), formulate the token-level Jensen-Shannon objective that aligns the two (\S\ref{sec:loss}), and detail how this objective is integrated with GRPO (\S\ref{sec:training}).
An overview is provided in Figure~\ref{fig:method}.

\subsection{Preliminaries}
\label{sec:prelim}

\paragraph{Trajectory Format.}
Following recent work on search-augmented reasoning~\citep{jin2025searchr1, shi2025autorefine}, the policy $\pi_\theta$ generates structured trajectories by interleaving reasoning with retrieval.
A trajectory $\tau$ consists of typed spans: \textsc{think} for internal reasoning, \textsc{search} for issuing a query to a fixed retriever $\mathcal{E}$, \textsc{documents} for the retrieved passages, and \textsc{answer} for the final prediction.
We write $\mathcal{A}_\tau$ for the set of token positions generated by the policy, that is, all positions inside think, search, and answer spans; document tokens are excluded as they are produced by the retriever. 
We further write $\mathcal{Q}_\tau \subseteq \mathcal{A}_\tau$ for the subset of positions that fall strictly between the \texttt{<search>} and \texttt{</search>} tags (i.e., search query tokens), which our self-distillation objective will target in \S\ref{sec:loss}.

\paragraph{GRPO with Trajectory-Level Reward.}
We adopt GRPO~\citep{shao2024deepseekmath} for policy optimization.
For each question $q$, a group of trajectories $\mathcal{G} = \{\tau_1, \ldots, \tau_G\}$ is sampled from the policy $\pi_{\theta_{\text{old}}}$, and this group will later serve as the hindsight source in \S\ref{sec:hindsight}.
Each trajectory receives a scalar reward $R_i$ based on the F1 score between the predicted and gold answers.
The group-normalized advantage shared by all tokens in $\tau_i$ is
\begin{equation}
\scriptsize
\hat{A}_i \;=\; \frac{R_i - \mathrm{mean}(\{R_j\}_{j=1}^G)}{\mathrm{std}(\{R_j\}_{j=1}^G)}.
\label{eq:grpo_adv}
\end{equation}

The policy is optimized with the standard GRPO loss
\begin{equation}
\scriptsize
\mathcal{L}_{\text{GRPO}}(\theta) \;=\; -\,\mathbb{E}\!\left[\min\!\left(r_p(\theta)\,\hat{A}_i,\ \mathrm{clip}(r_p(\theta),\,1-\epsilon,\,1+\epsilon)\,\hat{A}_i\right)\right], 
\quad r_p(\theta) = \frac{\pi_\theta(a_p \mid \tau_{<p})}{\pi_{\theta_{\text{old}}}(a_p \mid \tau_{<p})},
\label{eq:ppo}
\end{equation}
where $r_p(\theta)$ is the importance ratio.
Since this reward is trajectory-level, \emph{every token in $\tau_i$ is weighted by the same advantage $\hat{A}_i$ when updating the policy}, regardless of whether it belongs to a well-formed search query or a redundant one.

\subsection{Self-Distillation under Hindsight Conditioning}
\label{sec:hindsight}

As outlined in \S\ref{sec:intro}, SD-Search uses the same policy under two conditioning contexts: a \emph{student view} that contains only the inference-time context, and a \emph{teacher view} that additionally conditions on a hindsight information block as shown in Figure~\ref{fig:method}. 
We now make these two views precise.

\paragraph{Student View.}
The student is the policy under its standard inference-time conditioning.
For any action token at position $p \in \mathcal{A}_\tau$, the student distribution is
\begin{equation}
\scriptsize
P^{\text{stu}}_p \;=\; \pi_\theta(\cdot \mid \tau_{<p}),
\end{equation}
where $\tau_{<p}$ denotes the trajectory tokens preceding position $p$.

\paragraph{Teacher View.}
The teacher operates on the same parameters $\theta$ but on a context augmented with a \emph{hindsight block} $h(\mathcal{G})$. We write the resulting context abstractly as
\begin{equation}
\scriptsize
c^{\text{tch}}(\tau_i; \mathcal{G}) \;=\; \big[\,h(\mathcal{G});\; \tau_i\,\big],
\qquad
h(\mathcal{G}) \;=\; \{\mathcal{M}(\tau_j),\; y(\tau_j)\}_{j=1}^{G},
\label{eq:sd_teacher_ctx}
\end{equation}
where $[\,\cdot\,;\,\cdot\,]$ denotes that $h(\mathcal{G})$ is causally placed before the action positions of $\tau_i$ that we supervise; the concrete token-level layout, including where the block sits relative to the question prefix, is given in Appendix~\ref{app:teacher_input}.
Concretely, the hindsight block aggregates all rollouts that GRPO already samples for the question. For each $\tau_j$, it contains a masked view $\mathcal{M}(\tau_j)$ that retains only the search spans of $\tau_j$, together with a binary outcome label $y(\tau_j) \in \{\textsc{Correct}, \textsc{Incorrect}\}$. The label is derived by thresholding the F1 score of $\tau_j$'s final answer at $\rho$, which we set to $0$ by default so that any non-empty overlap with the gold answer counts as \textsc{Correct}. 
For any action token at position $p$ in $\tau_i$, the teacher distribution at the corresponding position is 
\begin{equation}
\scriptsize
P^{\text{tch}}_p \;=\; \mathrm{sg}\!\left[\,\pi_\theta\!\left(\cdot \,\middle|\, h(\mathcal{G}),\; \tau_{i,<p}\right)\right], 
\end{equation}
where $\tau_{i,<p}$ denotes the trajectory tokens of $\tau_i$ before position $p$, and $\mathrm{sg}[\cdot]$ indicates that $P^{\text{tch}}_p$ is treated as a constant target when computing gradients for the student.
The teacher forward over $c^{\text{tch}}(\tau_i; \mathcal{G})$ yields the teacher distributions $\{P^{\text{tch}}_p\}_{p \in \mathcal{Q}_\tau}$ that the self-distillation objective in \S\ref{sec:loss} aligns the student to, with $h(\mathcal{G})$ causally visible at every supervised position. 
At inference time, where the hindsight is unavailable, the two roles collapse back into a single policy without teacher forward pass.

\paragraph{Future Masking.}
We define the masking operator $\mathcal{M}$ as one that retains only the search spans of a trajectory and discards all other span types (\textsc{think}, \textsc{documents}, \textsc{answer}). 
Since the dropped spans are downstream of each search within the trajectory, we refer to this procedure as \emph{future masking}. 
Retaining the full trajectory in the hindsight block leaks the answer.
Once the teacher conditions on the retrieved \textsc{documents} and downstream \textsc{think} spans, the gold answer may become directly extractable from the hindsight prefix, and the teacher's search-position distributions collapse onto retrieval-skipping continuations, pushing the student to issue fewer searches rather than better ones. 
The masking operator $\mathcal{M}$ prevents this collapse; we quantify its severity in \S\ref{sec:ablation_teacher}.

\paragraph{Outcome Conditioning.}
Attaching the outcome label $y(\tau_j)$ to each rollout in the hindsight information turns the conditioning into a contrast.
Conditioned on a \textsc{Correct} label, the teacher tends to reaffirm the query tokens that led to success; conditioned on an \textsc{Incorrect} label, it redistributes mass away from the failed queries and toward query patterns shared with the successful siblings.
The student, which sees the same $\tau_i$ prefix without any hindsight, inherits both effects by matching the teacher through the divergence objective introduced in \S\ref{sec:loss}.
In particular, when $j = i$, the hindsight block exposes $\tau_i$'s own search spans to the teacher, raising a natural concern that the teacher may reduce to copying $\tau_i$'s queries through the prepended block.
We address this concern in \S\ref{sec:ablation_teacher} with a leave-one-out variant that excludes $\tau_i$ from $h(\mathcal{G})$ while keeping its $G-1$ sibling rollouts.

\subsection{Token-Level Jensen-Shannon Objective}
\label{sec:loss}

For each trajectory $\tau$, the self-distillation loss averages a token-level divergence between the teacher and student distributions:
\begin{equation}
\scriptsize
\mathcal{L}_{\text{SD}}(\tau) \;=\; \frac{1}{|\mathcal{Q}_\tau|} \sum_{p \in \mathcal{Q}_\tau} \mathrm{JSD}\!\left(P^{\text{tch}}_p \,\Vert\, P^{\text{stu}}_p\right),
\label{eq:sd_loss}
\end{equation}
where the Jensen-Shannon divergence is the symmetric form
\begin{equation}
\scriptsize
\mathrm{JSD}(P \Vert Q) \;=\; \tfrac{1}{2}\mathrm{KL}\!\left(P \,\Big\Vert\, \tfrac{P+Q}{2}\right) + \tfrac{1}{2}\mathrm{KL}\!\left(Q \,\Big\Vert\, \tfrac{P+Q}{2}\right).
\end{equation}
We choose JSD~\citep{Agarwal2024jsdloss} over forward or reverse KL for two reasons.
It is symmetric, thus avoiding biasing the student toward either mode-seeking or mode-covering behavior, and it is bounded by $\log 2$, which empirically improves training stability when teacher and student distributions occasionally diverge sharply on early-training rollouts.
For computational efficiency, we additionally restrict both distributions to the union of their top-$k$ tokens (renormalizing to sum to one over this support), with $k = 50$ by default, and we ablate this choice in \S\ref{sec:ablation_hparams}.
We restrict the sum to $\mathcal{Q}_\tau$ because these positions directly express each search decision, whereas the trajectory-level reward already propagates, albeit sparsely, to all action positions; we ablate broadening this scope to $\mathcal{A}_\tau$ in \S\ref{sec:ablation_objective}.
For trajectories with $|\mathcal{Q}_\tau|=0$, we set $\mathcal{L}_{\mathrm{SD}}(\tau)=0$ and train only with the GRPO term.

\subsection{Training Procedure}
\label{sec:training}

\paragraph{Total Objective.}
The self-distillation loss is added to the standard GRPO objective without modifying its advantage estimator:
\begin{equation}
\scriptsize
\mathcal{L}_{\text{total}} \;=\; \mathcal{L}_{\text{GRPO}} \;+\; \alpha_{\text{SD}}\, \mathcal{L}_{\text{SD}}.
\label{eq:sd_total_loss}
\end{equation}
These two terms are complementary.
$\mathcal{L}_{\text{GRPO}}$ provides coarse trajectory-level reinforcement through the advantage $\hat{A}_i$, modulating token logits indirectly via reward.
$\mathcal{L}_{\text{SD}}$ provides dense token-level supervision through direct distribution matching.
Because $\mathcal{L}_{\text{SD}}$ never enters the advantage computation, existing infrastructure can adopt SD-Search by adding a single auxiliary forward pass and an extra term to the loss, with no changes to rollout sampling, reward shaping, or advantage normalization.

\paragraph{Warmup.}
At the start of training, the policy has not yet learned to follow the structured trajectory format, and its rollouts contain few well-formed search queries.
Distilling from such trajectories injects noise rather than signal.
We therefore apply a warmup of $T_{\text{warm}}$ training steps (50 by default) during which $\alpha_{\text{SD}}$ is held at zero; afterwards it is set to its target value ($10^{-3}$ by default).

\paragraph{Computational Cost.}
SD-Search adds a training-only teacher evaluation over the hindsight-augmented context and a JSD computation at search-query positions.
Because the teacher target is stop-gradiented, the additional backward pass comes only from the student-side distillation loss, not from updating a separate teacher model.
Inference cost is unchanged relative to the outcome-reward baseline.
Empirically, a full 200-step run at the 3B scale takes about 11.9 hours on 8$\times$H800, compared with 10.3 hours for the AutoRefine outcome-reward baseline under the same hardware, batch size, rollout configuration, and retriever, corresponding to a $15.5\%$ end-to-end wall-clock overhead.
This overhead remains within the standard RL loop: SD-Search requires no external teacher inference, unlike methods that synthesize training trajectories with a larger model, and no auxiliary API annotation phase, unlike methods that rely on GPT-4o-generated sub-question annotations.
A per-stage cost breakdown is provided in Appendix~\ref{app:cost_breakdown}.

\section{Experiments}
\label{sec:exp}

\subsection{Experimental Setup}
\label{sec:setup}

\paragraph{Datasets.}
\par
Following~\citep{jin2025searchr1, shi2025autorefine}, we train on a combined NQ and HotpotQA training set and evaluate on seven benchmarks: three single-hop (NQ~\citep{kwiatkowski2019naturalquestions}, TriviaQA~\citep{joshi2017triviaqa}, PopQA~\citep{mallen2023popqa}) and four multi-hop (HotpotQA~\citep{yang2018hotpotqa}, 2Wiki~\citep{ho20202wiki}, MuSiQue~\citep{trivedi2022musique}, Bamboogle~\citep{press2023bamboogle}). 
We report Exact Match (EM) accuracy after standard normalization. 
Dataset sizes and split statistics are reported in Appendix~\ref{app:datasets}.

\paragraph{Baselines.}
We compare SD-Search against three groups: retrieval-free baselines (Direct Generation, SFT, R1~\citep{deepseek2025r1}), a single-round retrieval baseline (Naive RAG~\citep{lewis2020naiverag}), and multi-round search-augmented reasoning methods. 
The last group further splits into inference-time prompting methods (Search-o1~\citep{li2025searcho1}, IRCoT~\citep{trivedi2023ircot}), outcome-reward RL methods (Search-R1~\citep{jin2025searchr1}, ReSearch~\citep{chen2025research}, AutoRefine~\citep{shi2025autorefine}, MR-Search~\citep{xiao2025mrsearch}), and process-supervision methods that rely on external resources (StepSearch~\citep{wang2025stepsearch} uses GPT-4o sub-question annotations, and Thinker~\citep{xu2025thinker} builds its training data with a 72B teacher).
All multi-round baselines share the same fixed retrieval corpora (December 2018 Wikipedia), fixed retriever (E5-base-v2~\citep{wang2024e5}), and retrieval depth of three passages per query. 
Baseline numbers are taken from their original papers when evaluation protocols match.

\paragraph{Implementation.}
SD-Search is implemented based on the veRL framework~\citep{Sheng2025hybridflow} with GRPO as the policy optimizer. 
Main results use Qwen2.5-3B and Qwen2.5-7B (both base and instruct) as backbones; all ablations are conducted with Qwen2.5-3B-Base.
SD-Search introduces four additional hyperparameters, set to $\alpha_{\text{SD}}{=}10^{-3}$, $T_{\text{warm}}{=}50$, top-$k{=}50$, and $\rho{=}0$ throughout all experiments without per-dataset tuning. 
GRPO hyperparameters, training-time resources, and the full configuration are listed in Appendix~\ref{app:hparams}.

\subsection{Main Results}
\label{sec:main_results}

Table~\ref{tab:main} reports EM on all seven benchmarks at the 3B scale. SD-Search-Base reaches $0.428$ average EM, improving over every outcome-reward baseline and statistically matching the leading process-supervision baseline Thinker-Instruct (seed-level variance is reported in Appendix~\ref{app:statistical}), despite using no external teacher model.

\begin{table*}[!htbp]
\centering
\caption{EM on seven QA benchmarks with \textbf{Qwen2.5-3B}. Dashes denote settings not reported by the original paper. Bold denotes the best and underline the second best per column.}
\label{tab:main}
\scriptsize
\setlength{\tabcolsep}{2.75pt}
\begin{tabular}{l cc ccc cccc c}
\toprule
\multirow{2}{*}[-0.8ex]{\textbf{Method}} & \multirow{2}{*}[-0.8ex]{\makecell[c]{\textbf{Training} \\ \textbf{Pipeline}}} & \multirow{2}{*}[-0.8ex]{\makecell[c]{\textbf{External} \\ \textbf{Teacher}}} & \multicolumn{3}{c}{\textbf{Single-Hop}} & \multicolumn{4}{c}{\textbf{Multi-Hop}} & \multirow{2}{*}[-0.8ex]{\textbf{Avg}.} \\
\cmidrule(lr){4-6}\cmidrule(lr){7-10}
~ & ~ & ~ & NQ & TriviaQA & PopQA & HotpotQA & 2Wiki & MuSiQue & Bamboogle & ~ \\
\midrule
\rowcolor{gray!10}
\multicolumn{11}{c}{\textit{\textbf{w/o Retrieval}}} \\
Direct Generation         & Zero-Shot & -- & 0.106 & 0.288 & 0.108 & 0.149 & 0.244 & 0.020 & 0.024 & 0.134 \\
SFT                       & SFT & -- & 0.249 & 0.292 & 0.104 & 0.186 & 0.248 & 0.044 & 0.112 & 0.176 \\
R1-Instruct               & Pure RL & -- & 0.210 & 0.449 & 0.171 & 0.208 & 0.275 & 0.060 & 0.192 & 0.224 \\
R1-Base                   & Pure RL & -- & 0.226 & 0.455 & 0.173 & 0.201 & 0.268 & 0.055 & 0.224 & 0.229 \\
\midrule
\rowcolor{gray!10}
\multicolumn{11}{c}{\textit{\textbf{w/ Single-Hop Retrieval}}} \\
Naive RAG                 & Zero-Shot & -- & 0.348 & 0.544 & 0.387 & 0.255 & 0.226 & 0.047 & 0.080 & 0.270 \\
\midrule
\rowcolor{gray!10}
\multicolumn{11}{c}{\textit{\textbf{w/ Multi-Hop Retrieval}}} \\
Search-o1                 & Prompting & -- & 0.238 & 0.472 & 0.262 & 0.221 & 0.218 & 0.054 & 0.320 & 0.255 \\
IRCoT                     & Prompting & -- & 0.111 & 0.312 & 0.200 & 0.164 & 0.171 & 0.067 & 0.240 & 0.181 \\
ReSearch-Base             & Pure RL & -- & 0.427 & 0.597 & 0.430 & 0.305 & 0.272 & 0.074 & 0.128 & 0.319 \\
ReSearch-Instruct         & Pure RL & -- & 0.365 & 0.571 & 0.395 & 0.351 & 0.272 & 0.095 & 0.266 & 0.331 \\
Search-R1-Base            & Pure RL & -- & 0.421 & 0.583 & 0.413 & 0.297 & 0.274 & 0.066 & 0.128 & 0.312 \\
Search-R1-Instruct        & Pure RL & -- & 0.397 & 0.565 & 0.391 & 0.331 & 0.310 & 0.124 & 0.232 & 0.336 \\
AutoRefine-Base           & Pure RL & -- & 0.467 & 0.620 & 0.450 & 0.405 & 0.393 & 0.157 & 0.344 & 0.405 \\
AutoRefine-Instruct       & Pure RL & -- & 0.436 & 0.597 & 0.447 & 0.404 & 0.380 & 0.169 & 0.336 & 0.396 \\
MR-Search-Base            & Pure RL & -- & \textbf{0.477} & \textbf{0.635} & 0.460 & 0.419 & 0.401 & 0.165 & 0.344 & 0.414 \\
StepSearch-Base           & Pure RL & GPT-4o & --    & --    & --    & 0.329 & 0.339 & 0.181 & 0.328 & --    \\
StepSearch-Instruct       & Pure RL & GPT-4o & --    & --    & --    & 0.345 & 0.320 & 0.174 & 0.344 & --    \\
Thinker-Instruct          & Synthesize $\rightarrow$ SFT & Qwen2.5-72B & 0.439 & 0.598 & \textbf{0.469} & 0.400 & \textbf{0.469} & \textbf{0.214} & \textbf{0.424} & \textbf{0.430} \\
\rowcolor{cyan!10}
\textbf{SD-Search-Base}     & Pure RL & -- & \underline{0.470} & \underline{0.624} & \underline{0.467} & \textbf{0.425} & \underline{0.420} & \underline{0.188} & 0.402 & \underline{0.428} \\
\rowcolor{cyan!10}
\textbf{SD-Search-Instruct} & Pure RL & -- & 0.469 & \underline{0.624} & 0.465 & \underline{0.424} & 0.419 & 0.184 & \underline{0.404} & 0.427 \\
\bottomrule
\end{tabular}
\end{table*}

\par
Against \textbf{pure outcome-reward methods}, SD-Search-Base surpasses AutoRefine-Base by $2.3$ points on average and MR-Search-Base by $1.4$ points. 
The gains are larger on benchmarks requiring multiple retrieval hops (Bamboogle $+5.8$, MuSiQue $+3.1$, 2Wiki $+2.7$ over AutoRefine-Base).
The self-distillation loss is applied at every search-query token, so trajectories with more search steps receive proportionally denser supervision. 
Multi-hop questions also induce greater within-trajectory variance in query quality (some hops route correctly, others retrieve irrelevant passages), and the token-level JSD signal explicitly distinguishes these positions, whereas a trajectory-level advantage averages over them. The two effects compound, accounting for the larger gains on multi-hop benchmarks.
SD-Search-Instruct tracks the Base variant closely at $0.427$, indicating the benefit is not tied to any particular instruction-tuning prior.
Against \textbf{process-supervision methods}, SD-Search-Base is statistically indistinguishable from Thinker on average ($0.428$ versus $0.430$), while using only the question-answer pairs required by standard GRPO, without the 72B teacher or supervised distillation stage that Thinker depends on. 
It also surpasses StepSearch on every multi-hop benchmark, with gains of $+9.6$ points on HotpotQA and $+8.1$ on 2Wiki over StepSearch-Base, while not requiring GPT-4o-generated sub-question annotations that StepSearch relies on.
These results indicate that: \emph{the step-level signal that process-supervision methods obtain from external sources can be recovered from the policy itself through hindsight conditioning, without compromising performance.}
Training dynamics further support this interpretation, with SD-Search's gain driven by query quality rather than search volume (Appendix~\ref{app:dynamics}).
An illustrative rollout comparison on a 2Wiki question is provided in Appendix~\ref{app:case}.

\par
Table~\ref{tab:main_7b} reports the same comparison at the 7B scale.
SD-Search-Instruct reaches $0.476$ average EM, improving over MR-Search-Base by $1.6$ points and AutoRefine-Base by $2.1$ points. 
It also surpasses Thinker-Instruct ($0.452$) by $2.4$ points while using no external teacher model, confirming at this scale the same pattern observed at 3B.
The per-benchmark gains are concentrated on multi-hop settings (HotpotQA $+2.0$, MuSiQue $+3.5$ over AutoRefine-Base), and SD-Search-Base reaches $0.471$, within $0.5$ point of the Instruct variant, indicating that the hindsight signal transfers to both pretraining-only and instruction-tuned starting points.
Notably, while AutoRefine-Base improves by $+5.0$ points from 3B to 7B ($0.405\to0.455$), Thinker-Instruct only adds $+2.2$ points ($0.430\to0.452$), lagging behind its own non-distilled outcome-reward baseline.
By contrast, SD-Search scales with the student under both conditioning regimes: comparing Base-to-Base, SD-Search-Base gains $+4.3$ points ($0.428\to0.471$), tracking AutoRefine-Base's $+5.0$ within $0.7$ point; comparing Instruct-to-Instruct, SD-Search-Instruct gains $+4.9$ points ($0.427\to0.476$), more than double Thinker-Instruct's $+2.2$. 
This is consistent with the view that on-policy hindsight self-distillation scales with the student's own in-context reading capacity rather than against a fixed external reference.
An extended scaling analysis at 1.5B, 3B, 7B, and 14B (Appendix~\ref{app:scaling}) confirms this pattern: 
Thinker's gain over AutoRefine shrinks monotonically with scale and turns negative from 7B onward, while SD-Search's gain remains positive at every scale, narrowing at 14B but never crossing zero.

\begin{table}[!htbp]
\centering
\caption{EM on seven QA benchmarks with \textbf{Qwen2.5-7B}. Dashes denote settings not reported by the original paper. Bold denotes the best and underline the second best per column.}
\label{tab:main_7b}
\scriptsize
\setlength{\tabcolsep}{2.75pt}
\begin{tabular}{l cc ccc cccc c}
\toprule
\multirow{2}{*}[-0.8ex]{\textbf{Method}} & \multirow{2}{*}[-0.8ex]{\makecell[c]{\textbf{Training} \\ \textbf{Pipeline}}} & \multirow{2}{*}[-0.8ex]{\makecell[c]{\textbf{External} \\ \textbf{Teacher}}} & \multicolumn{3}{c}{\textbf{Single-Hop}} & \multicolumn{4}{c}{\textbf{Multi-Hop}} & \multirow{2}{*}[-0.8ex]{\textbf{Avg}.} \\
\cmidrule(lr){4-6}\cmidrule(lr){7-10}
~ & ~ & ~ & NQ & TriviaQA & PopQA & HotpotQA & 2Wiki & MuSiQue & Bamboogle & ~ \\
\midrule
Search-R1-Base          & Pure RL & -- & 0.469 & 0.627 & 0.449 & 0.410 & 0.272 & 0.173 & 0.456 & 0.408 \\
Search-R1-Instruct      & Pure RL & -- & 0.393 & 0.610 & 0.397 & 0.370 & 0.414 & 0.146 & 0.368 & 0.385 \\
AutoRefine-Base         & Pure RL & -- & 0.484 & 0.659 & 0.487 & 0.451 & 0.405 & 0.187 & 0.512 & 0.455 \\
MR-Search-Base          & Pure RL & -- & \textbf{0.502} & \underline{0.666} & 0.472 & \underline{0.468} & 0.436 & 0.221 & 0.452 & 0.460 \\
StepSearch-Base         & Pure RL & GPT-4o & --    & --    & --    & 0.380 & 0.385 & 0.216 & 0.467 & --    \\
StepSearch-Instruct     & Pure RL & GPT-4o & --    & --    & --    & 0.386 & 0.366 & \textbf{0.226} & 0.400 & --    \\
Thinker-Instruct        & Synthesize $\rightarrow$ SFT & Qwen2.5-72B & 0.450 & 0.642 & 0.484 & 0.421 & \textbf{0.469} & 0.221 & 0.480 & 0.452 \\
\rowcolor{cyan!10}
\textbf{SD-Search-Base}     & Pure RL & -- & \underline{0.500} & 0.665 & \underline{0.489} & \underline{0.468} & 0.438 & 0.204 & \underline{0.532} & \underline{0.471} \\
\rowcolor{cyan!10}
\textbf{SD-Search-Instruct} & Pure RL & -- & 0.495 & \textbf{0.668} & \textbf{0.494} & \textbf{0.471} & \underline{0.441} & \underline{0.222} & \textbf{0.544} & \textbf{0.476} \\
\bottomrule
\end{tabular}
\end{table}

\subsection{Ablation Studies}
\label{sec:ablation}

We run three families of ablations on Qwen2.5-3B-Base. 
The first (\S\ref{sec:ablation_teacher}) isolates the construction of the hindsight block, asking what information the teacher should and should not see. 
The second (\S\ref{sec:ablation_objective}) isolates the student-teacher alignment objective itself. 
The third (\S\ref{sec:ablation_hparams}) reports sensitivity to the four SD-Search hyperparameters around their defaults.

\subsubsection{Hindsight Block Construction}
\label{sec:ablation_teacher}

\begin{figure}[!htbp]
\centering
\includegraphics[width=0.75\linewidth]{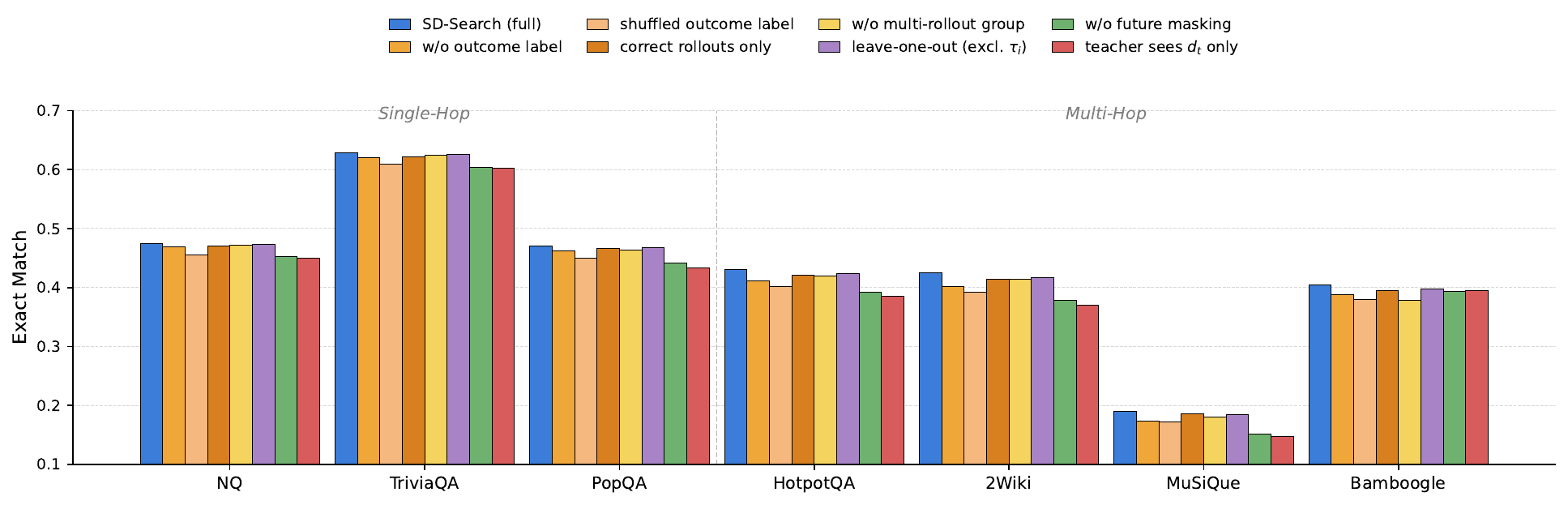}
\caption{Hindsight block construction ablation on Qwen2.5-3B-Base. Eight
configurations cluster into three colour-coded families. The
\emph{outcome-label} family (three orange shades) dissects the role of
outcome labels. The \emph{group-structure} family (yellow, purple)
probes which part of the rollout group carries the signal. The
\emph{future-content} family (green, red) probes whether downstream
tokens leak the answer. Drops are concentrated on multi-hop benchmarks
for every family.}
\label{fig:ablation_hindsight}
\end{figure}

The hindsight block $h(\mathcal{G})$ makes three design choices: attaching
an outcome label to each rollout, masking non-search tokens, and
aggregating over the entire rollout group. 
We modify each choice in turn
while keeping the rest of the pipeline fixed, and additionally probe
whether the outcome label is used as a \textsc{Correct}/\textsc{Incorrect}
contrast or as label conditioning (by shuffling labels or filtering to
correct rollouts only), and whether the block can be replaced by the
current-step documents $d_t$ alone. Figure~\ref{fig:ablation_hindsight}
reports EM under each configuration.

\par
The outcome-label family (three orange shades in
Figure~\ref{fig:ablation_hindsight}) probes the labels themselves.
Removing the outcome label costs $1.4$ points, since the teacher cannot tell queries from successful rollouts apart from those from failed ones. Shuffling each label uniformly at random costs $2.3$ points, \emph{worse} than removing them: random labels are actively misleading, isolating the \textsc{Correct}/\textsc{Incorrect} contrast
(rather than label conditioning) as the source of the gain. Keeping only successful skeletons (correct rollouts only) costs $0.7$ points: one-sided imitation recovers part but not all of the contrast benefit.
 
\par
Removing the multi-rollout group costs $1.0$ points; the leave-one-out variant, which removes $\tau_i$ from $h(\mathcal{G})$ while keeping its $G{-}1$ siblings, is its complement and costs only $0.5$ points.
The cross-rollout contrast contributes more than $\tau_i$'s own search spans, and the teacher cannot be trivially reduced to a copy of $\tau_i$ through the prepended block.
 
\par
Removing future masking costs $3.0$ points on average, with the sharpest losses on multi-hop benchmarks (HotpotQA $-3.8$, 2Wiki $-4.7$). 
Once the teacher conditions on downstream think tokens and retrieved documents, the gold answer becomes predictable from the prefix and the teacher's distribution concentrates on retrieval-skipping continuations.
Replacing the entire block with the current-step documents $d_t$ alone produces the largest drop of $3.4$ points: the teacher either skips the search or echoes entities from $d_t$ rather than assessing query quality. 
Local retrieval traces therefore do not recover the role of the full hindsight block as a query-quality supervisor.

\subsubsection{Student-Teacher Alignment Objective}
\label{sec:ablation_objective}

\par
The self-distillation loss in Eq.~\ref{eq:sd_loss} makes two choices: Jensen-Shannon divergence as the distributional distance, and a summation scope restricted to the search-query positions $\mathcal{Q}_\tau$. 
We probe each choice by replacing it with alternatives while keeping the rest fixed. 
Results are reported in Table~\ref{tab:ablation_objective}.

\begin{table}[!htbp]
\centering
\caption{Alignment objective ablation on Qwen2.5-3B-Base. The top block varies the divergence function while keeping the scope fixed to $\mathcal{Q}_\tau$; the bottom row broadens the scope from $\mathcal{Q}_\tau$ to $\mathcal{A}_\tau$ while keeping the divergence fixed to JSD.}
\label{tab:ablation_objective}
\scriptsize
\setlength{\tabcolsep}{3.5pt}
\begin{tabular}{l ccc cccc c}
\toprule
\multirow{2}{*}{\textbf{Configuration}} & \multicolumn{3}{c}{\textbf{Single-Hop}} & \multicolumn{4}{c}{\textbf{Multi-Hop}} & \multirow{2}{*}{\textbf{Avg}.} \\
\cmidrule(lr){2-4}\cmidrule(lr){5-8}
~ & NQ & TriviaQA & PopQA & HotpotQA & 2Wiki & MuSiQue & Bamboogle & ~ \\
\midrule
SD-Search-Base (full)  & 0.470 & 0.624 & 0.467 & 0.425 & 0.420 & 0.188 & 0.402 & \textbf{0.428} \\
\midrule
\; forward KL instead of JSD                      & 0.465 & 0.619 & 0.459 & 0.413 & 0.405 & 0.178 & 0.387 & 0.418 \\
\; reverse KL instead of JSD                      & 0.447 & 0.619 & 0.464 & 0.408 & 0.430 & 0.153 & 0.376 & 0.414 \\
\; MSE instead of JSD                             & 0.460 & 0.614 & 0.454 & 0.400 & 0.393 & 0.168 & 0.362 & 0.407 \\
\midrule
\; scope $\mathcal{A}_\tau$ instead of $\mathcal{Q}_\tau$ & 0.466 & 0.620 & 0.461 & 0.417 & 0.410 & 0.181 & 0.392 & 0.421 \\
\bottomrule
\end{tabular}
\end{table}

\begin{figure}[!htbp]
\centering
\includegraphics[width=\linewidth]{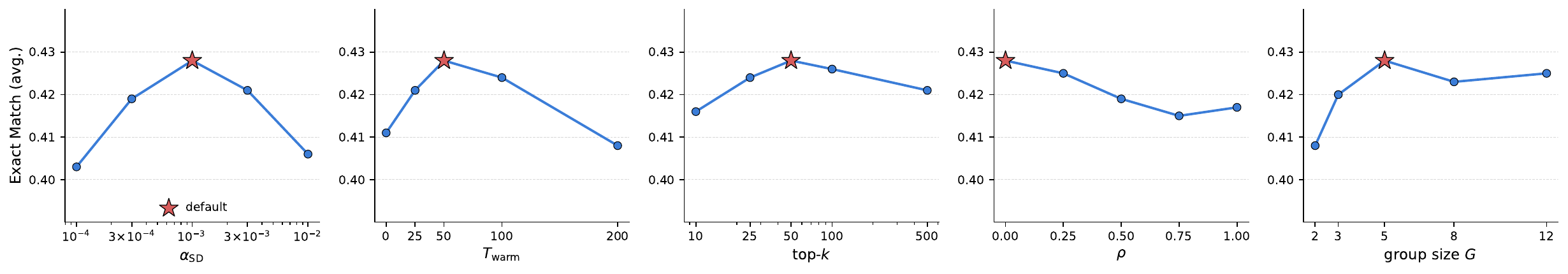}
\caption{Sensitivity of hyperparameters on Qwen2.5-3B-Base. Each panel
varies one hyperparameter while the others are fixed at their defaults
(marked with a red star). The first four panels sweep the
self-distillation hyperparameters; the fifth sweeps the GRPO group
size $G$.}
\label{fig:sensitivity}
\end{figure}

\par
All three alternatives underperform JSD, by $1.0$, $1.4$, and $2.1$ points on average for forward KL~\citep{hinton2015distill}, reverse KL~\citep{Gu2024minillm}, and MSE respectively. 
The two asymmetric KLs underperform in opposite ways. 
\textbf{Forward KL} is mode-covering: it requires the student to place non-zero mass on \emph{every} token supported by the teacher, preventing the student from committing sharply to any single keyword even when the teacher itself is highly concentrated. The resulting entropy in the student's predictions slightly blunts the signal.
\textbf{Reverse KL} is mode-seeking, collapsing the student onto the teacher's single highest-probability token and foreclosing exploration, a cost that shows up most on multi-hop benchmarks where multiple near-equivalent query reformulations are often viable.
\textbf{MSE} (applied to the raw logits) performs worst because it treats every logit coordinate as equally important, ignoring the probability-simplex structure that the KL family respects; small logit shifts on irrelevant tokens are penalized as heavily as shifts on the high-probability tokens that actually carry the query information.
These results support the symmetry and boundedness properties we attributed to JSD in \S\ref{sec:loss}.
\textbf{Broadening the scope} from $\mathcal{Q}_\tau$ to all policy-generated positions $\mathcal{A}_\tau$ costs $0.7$ points, a smaller effect than any of the divergence swaps. We believe this reflects a subtler failure mode: at think and answer positions, the teacher's distribution is partially shaped by the outcome labels in $h(\mathcal{G})$, which pushes it toward sharply concentrated continuations that short-circuit the reasoning chain. Aligning the student to these overly confident targets suppresses the exploratory think patterns that the trajectory-level advantage was relying on.

\subsubsection{Hyperparameter Sensitivity}
\label{sec:ablation_hparams}

We sweep each of the four SD-Search hyperparameters together with the GRPO group size $G$ in turn while fixing the others at their defaults.
Figure~\ref{fig:sensitivity} reports EM across the seven benchmarks.
All five defaults, including the group size $G{=}5$, sit at the peak of their respective curves, and the overall amplitude across the sweep window is modest, never exceeding $2.5$ points.

\par
$\alpha_{\text{SD}}$ shows a mild inverted-U shape around its default: too small a coefficient underweights the step-level signal, while too large a coefficient overwhelms the GRPO exploration that the advantage term was supposed to induce.
$T_{\text{warm}}$ shows a similar shape but for a different reason: starting distillation too early feeds noise from poorly formed trajectories into the student, while starting it too late forfeits the training window in which query formulation is most plastic.
The top-$k$ sweep is nearly flat in the range $k \in [25, 100]$, with modest degradation at both ends. This insensitivity is consistent with our observation in \S\ref{sec:loss} that most of the JSD mass at each position is concentrated on a small number of high-probability tokens.
$\rho$ peaks at the default $\rho{=}0.0$ and drops as the threshold tightens, reaching its minimum near $\rho{=}0.75$, where too many rollouts get relabeled to \textsc{Incorrect} and the hindsight block loses its outcome contrast. 
The small recovery at $\rho=1.0$ reflects that a strict perfect-match criterion successfully filters out ambiguous outcomes, leaving a sharper but sparser contrast between \textsc{Correct} and \textsc{Incorrect} trajectories.

\section{Limitations}
\label{sec:limitations}

SD-Search has two main limitations. 
First, the hindsight contrast relies on outcome labels $y(\tau_j)$ derived from gold answers, as it is inherited from GRPO itself, which restricts the method to tasks with reliably scorable references. 
Extending SD-Search to open-ended generation requires a substitute for the binary outcome label, for example a preference score from a learned model or a majority-vote proxy across the rollout group, which we leave to future work.
Second, although the multi-rollout group is designed to keep the hindsight signal informative when the current rollout $\tau_i$ fails, the outcome contrast in the hindsight block degenerates whenever the group becomes label-homogeneous in either direction. 
The all-failure scenario, in which every rollout in the group ends \textsc{Incorrect}, is the same scenario where GRPO's trajectory-level advantages also collapse; the symmetric all-success regime, in which every rollout ends \textsc{Correct}, removes the contrast as well, and we observe a corresponding narrowing of SD-Search's gain over AutoRefine at 14B (Appendix~\ref{app:scaling}), where higher base success rates make group-uniform \textsc{Correct} labels more frequent. 
Designing a hindsight construction that remains informative under both group-uniform scenarios is an open problem.

\section{Conclusion}
\label{sec:conclusion}

We introduce SD-Search, a reinforcement learning framework that supplies step-level supervision for search-augmented reasoning through on-policy hindsight self-distillation. 
The same policy serves as both student and teacher, with the teacher conditioned on a hindsight block that aggregates the rollout group and its outcomes. 
Aligning the student to the teacher on search-query positions yields a dense gradient signal that complements GRPO's trajectory-level reward, without any external teacher model or auxiliary annotations.
On seven QA benchmarks, SD-Search outperforms outcome-reward baselines and matches process-supervision methods that rely on much larger external systems, at a $15.5\%$ end-to-end training-time overhead which is entirely contained within the standard RL loop, without external teacher inference, annotation phase, or additional inference cost.
More broadly, the construction of a teacher by conditioning the same policy on hindsight observations points toward a more general recipe for extracting step-level signal from trajectory-level reinforcement learning, beyond the search-augmented reasoning setting studied here.
\bibliographystyle{plainnat}
\bibliography{reference}

\clearpage

\appendix

\section{Related Work}
\label{sec:related}

\paragraph{Search-augmented reasoning with reinforcement learning.}
A line of recent work trains LLMs to invoke retrieval tools during reasoning, optimizing the policy with outcome-level reinforcement learning. 
Search-R1~\citep{jin2025searchr1} introduces the search-during-think paradigm and uses GRPO to train the agent on final-answer correctness. 
ReSearch~\citep{chen2025research} and AutoRefine~\citep{shi2025autorefine} extend this formulation with stronger reward shaping and structured trajectories, and MR-Search~\citep{xiao2025mrsearch} further introduces cross-episode reflection. 
All of these methods share a common property: supervision operates only at the trajectory level, so every token within a rollout, regardless of whether a query is well or poorly formed, receives the same gradient signal. 
SD-Search inherits the trajectory format and GRPO outer loop from this line of work but adds a dense step-level signal on top, without modifying the advantage estimator.

\paragraph{Process supervision for LLM reasoning agents.}
Several recent methods target the same limitation SD-Search addresses, namely the absence of step-level credit assignment under trajectory-level rewards. 
Thinker~\citep{xu2025thinker} constructs sub-question decompositions using a 72B teacher and supervises the student through distillation on these trajectories. 
StepSearch~\citep{wang2025stepsearch} derives step-wise rewards from GPT-4o-generated sub-question annotations. 
GiGPO~\citep{feng2025gigpo} takes a different route, exploiting repeated environment states to build step-level advantage groups within a trajectory; while originally developed for embodied and web-interaction agents, it exemplifies a structural approach to step-level signal.
These methods deliver clear improvements, but each obtains its step-level signal from outside the policy, either by importing external supervision (Thinker, StepSearch) or by assuming structural regularities in the trajectory (GiGPO). 
SD-Search differs in deriving the signal entirely from the policy itself, through conditioning on hindsight rather than through any external or structural proxy.

\paragraph{Self-distillation and hindsight learning.}
Recent self-distillation work shows that a single language model can serve as its own teacher when conditioned on privileged information the student does not see. 
The underlying principle of learning from privileged information dates back to \citet{lopezpaz2016unifyingdistillationprivilegedinformation}; recent applications include mathematical reasoning~\citep{zhao2026opsd} and continual learning~\citep{shenfeld2026sdft}.
In all these settings the privileged information is drawn from fixed external sources such as ground-truth traces or expert demonstrations. 
SD-Search applies the same mechanism in a reinforcement learning setting, where the privileged information is instead produced by the policy itself during rollout, namely the hindsight block over the rollout group and its outcomes. 
The term ``hindsight'' here is borrowed from Hindsight Experience Replay~\citep{andrychowicz2018hindsight} and the more recent hindsight credit assignment line~\citep{anna2019hindsightcredit}, both of which operate at the reward level whereas SD-Search operates at the token distribution level.

\clearpage
\section{Robustness Across Random Seeds}
\label{app:statistical}

To verify that our improvements are robust to random initialization, we re-train SD-Search-Base, AutoRefine-Base, and Thinker-Instruct with \emph{five} random seeds on Qwen2.5-3B under otherwise identical settings. 
Table~\ref{tab:seeds} reports the mean and standard deviation across seeds on each benchmark. 
The per-benchmark variability is uneven and is dominated by two factors: the size of the evaluation split and the intrinsic stochasticity of RL training. 
Standard deviations are smallest on the large single-hop test sets (NQ, TriviaQA, PopQA, all containing $>$3k examples; std $\approx$ 0.006--0.010), larger on the multi-hop dev sets (HotpotQA, 2Wiki, MuSiQue; std $\approx$ 0.009--0.015), and largest on Bamboogle (std $\approx$ 0.022--0.028), whose 125-example test split inflates the seed variance relative to the others. 
The two RL methods (AutoRefine-Base and SD-Search-Base) exhibit comparable seed-level variance, while Thinker-Instruct, which is trained via SFT on synthetic trajectories, is slightly more stable, consistent with SFT generally producing lower seed-to-seed variance than RL.

\par
Across seeds, SD-Search-Base's average EM ($0.428 \pm 0.008$) lies above AutoRefine-Base's ($0.404 \pm 0.008$) on every seed we ran, while overlapping with Thinker-Instruct's ($0.429 \pm 0.007$). 
We describe these gaps as consistent rather than as the result of formal significance testing: five seeds is a modest sample under RL training variance, and the per-benchmark standard deviations on small test sets such as Bamboogle remain large enough that single-benchmark conclusions should be read with caution. 
However, the pattern across the average column matches the main-text claim that SD-Search reliably beats outcome-reward baselines without an external teacher, and performs comparably to the external-teacher baseline.

\begin{table}[htbp]
\centering
\caption{Exact Match (mean$_{\text{std}}$) across five random seeds on Qwen2.5-3B.}
\label{tab:seeds}
\scriptsize
\setlength{\tabcolsep}{3pt}
\begin{tabular}{l ccc cccc c}
\toprule
\multirow{2}{*}{\textbf{Method}} & \multicolumn{3}{c}{\textbf{Single-Hop}} & \multicolumn{4}{c}{\textbf{Multi-Hop}} & \multirow{2}{*}{\textbf{Avg}}. \\
\cmidrule(lr){2-4}\cmidrule(lr){5-8}
 & NQ & TriviaQA & PopQA & HotpotQA & 2Wiki & MuSiQue & Bamboogle & \\
\midrule
AutoRefine-Base
  & $0.464_{\pm.009}$ & $0.621_{\pm.006}$ & $0.447_{\pm.008}$
  & $0.408_{\pm.013}$ & $0.390_{\pm.014}$ & $0.158_{\pm.012}$ & $0.339_{\pm.028}$
  & $0.404_{\pm.008}$ \\
Thinker-Instruct
  & $0.441_{\pm.008}$ & $0.594_{\pm.006}$ & $0.466_{\pm.007}$
  & $0.403_{\pm.011}$ & $0.466_{\pm.010}$ & $0.211_{\pm.009}$ & $0.421_{\pm.022}$
  & $0.429_{\pm.007}$ \\
\rowcolor{cyan!10}
SD-Search-Base
  & $0.470_{\pm.010}$ & $0.624_{\pm.007}$ & $0.467_{\pm.009}$
  & $0.425_{\pm.012}$ & $0.420_{\pm.015}$ & $0.188_{\pm.013}$ & $0.402_{\pm.026}$
  & $0.428_{\pm.008}$ \\
\bottomrule
\end{tabular}
\end{table}

\clearpage
\section{Dataset Statistics}
\label{app:datasets}

Table~\ref{tab:datasets} summarizes the seven question answering benchmarks used in our experiments. All datasets are sourced from the FlashRAG datasets collection. We construct the training set by combining the train splits of NQ and HotpotQA, following~\citet{jin2025searchr1}. For evaluation we use the test split where available (NQ, TriviaQA, PopQA, Bamboogle) and the dev split otherwise (HotpotQA, 2Wiki, MuSiQue).

\begin{table}[h]
\centering
\caption{Statistics of the seven benchmarks used in this paper.}
\label{tab:datasets}
\small
\setlength{\tabcolsep}{5pt}
\begin{tabular}{l ccc c}
\toprule
\multirow{2}{*}{\textbf{Dataset}} & \multicolumn{3}{c}{\textbf{Split sizes}} & \multirow{2}{*}{\textbf{Eval split}} \\
\cmidrule(lr){2-4}
 & Train & Dev & Test & \\
\midrule
NQ         & 79{,}168 &  8{,}757 &  3{,}610 & test \\
TriviaQA   & 78{,}785 &  8{,}837 & 11{,}313 & test \\
PopQA      & --       & --       & 14{,}267 & test \\
HotpotQA   & 90{,}447 &  7{,}405 & --       & dev  \\
2Wiki      & 15{,}000 & 12{,}576 & --       & dev  \\
MuSiQue    & 19{,}938 &  2{,}417 & --       & dev  \\
Bamboogle  & --       & --       &      125 & test \\
\bottomrule
\end{tabular}
\end{table}

\clearpage
\section{Hyperparameters}
\label{app:hparams}

Table~\ref{tab:hparams} lists the complete set of hyperparameters used in our experiments. The SD-Search-specific hyperparameters ($\alpha_{\text{SD}}, T_{\text{warm}}$, top-$k$, $\rho$) were selected on a validation subset of NQ and HotpotQA; their sensitivity is analyzed in \S\ref{sec:ablation_hparams}. All other values follow the default configuration of the veRL framework~\citep{Sheng2025hybridflow}.

\begin{table}[h]
\centering
\caption{Complete hyperparameter configuration for SD-Search training.}
\label{tab:hparams}
\small
\begin{tabular}{l l}
\toprule
\textbf{Hyperparameter} & \textbf{Value} \\
\midrule
\rowcolor{gray!10}
\multicolumn{2}{c}{\textit{\textbf{Training configuration}}} \\
Group size $G$                     & 5 \\
Training batch size                & 256 \\
Actor learning rate                & $1 \times 10^{-6}$ \\
Total training steps               & 200 \\
KL coefficient $\beta$             & 0.001 \\
Clip ratio $\epsilon$              & 0.2 \\
Max search calls per rollout       & 3 \\
Rollout temperature                & 1.0 \\
\midrule
\rowcolor{gray!10}
\multicolumn{2}{c}{\textit{\textbf{Retrieval configuration}}} \\
Retriever                          & E5-base-v2 \\
Corpus                             & Wikipedia (Dec.~2018) \\
Top-$k$ documents per query        & 3 \\
\midrule
\rowcolor{gray!10}
\multicolumn{2}{c}{\textit{\textbf{SD-Search specific configuration}}} \\
Distillation coefficient $\alpha_{\text{SD}}$     & $1 \times 10^{-3}$ \\
Warmup steps $T_{\text{warm}}$                    & 50 \\
Distribution truncation top-$k$                   & 50 \\
Outcome threshold $\rho$                        & 0 \\
Hindsight max length                       & 1024 tokens \\
Future masking                                    & enabled \\
Alignment scope                                   & $\mathcal{Q}_\tau$ \\
\bottomrule
\end{tabular}
\end{table}

\clearpage
\section{Training Cost Breakdown}
\label{app:cost_breakdown}

We measure SD-Search's training cost both end-to-end (full 200-step run wall-clock) and at per-step granularity (a separate 100-step instrumented run). Both numbers are reported in Table~\ref{tab:cost_breakdown}; together they let us isolate which stages of the training loop the SD-Search additions actually touch.

\paragraph{End-to-end.}
On 8$\times$H800, the full 200-step training run takes about 11.9h with SD-Search enabled (Qwen2.5-3B-Base) versus about 10.3h for the AutoRefine baseline under identical hardware, batch size, rollout configuration, and retriever, a $+15.5\%$ end-to-end wall-clock overhead.

\paragraph{Per-step breakdown.} 
To attribute this overhead to specific training stages, we instrument both runs at the level of \texttt{update\_policy}'s internal sections (PPO forward, PPO backward, SD-Search compute, SD-Search backward, optimizer + communication) using CUDA-synced wall-clock timers, and average over a 100-step run skipping the first 50 steps as warmup. 
Table~\ref{tab:cost_breakdown} reports the result. 
Roughly $94\%$ of the per-step overhead is contributed by the SD-Search compute stage (teacher forward + JSD computation, $+22.0\%$ of baseline step time) and the SD-Search backward stage ($+17.5\%$). 
All other stages, including rollout generation, reward computation, PPO forward/backward, optimizer step, and communication, are unchanged within measurement noise ($|\Delta| \le 2.1\%$ of baseline; the rollout stage is in fact slightly faster for SD-Search at $-2.1\%$, attributable to run-to-run variance since the rollout pipeline is identical between the two methods), confirming that SD-Search does not affect the rollout, retrieval, advantage, or core PPO pipeline.

\begin{table}[htbp]
\centering
\caption{Per-step wall-clock breakdown on Qwen2.5-3B-Base, 8$\times$H800, batch=256, $G$=5. 
Means over 50 steps after warmup skip. The two SD-Search rows account for nearly all of the overhead; all other stages are unchanged within measurement noise.}
\label{tab:cost_breakdown}
\footnotesize
\begin{tabular}{l cccc}
\toprule
\textbf{Stage} & \textbf{AutoRefine} (s) & \textbf{SD-Search} (s) & $\Delta$ (s) & $\Delta$ (\% of baseline total) \\
\midrule
Rollout generation     & 66.16 & 64.23 & $-1.93$ & $-2.1\%$ \\
Reward + advantage     & 2.69  & 3.31  & $+0.61$ & $+0.7\%$ \\
Reference logprob      & 3.67  & 4.13  & $+0.46$ & $+0.5\%$ \\
PPO forward            & 5.70  & 6.01  & $+0.31$ & $+0.3\%$ \\
PPO backward           & 9.25  & 9.97  & $+0.72$ & $+0.8\%$ \\
SD-Search compute      & ---   & 20.31 & $+20.31$ & $+22.0\%$ \\
SD-Search backward     & ---   & 16.13 & $+16.13$ & $+17.5\%$ \\
Optimizer + comm       & 0.13  & 0.13  & $\sim 0$ & $\sim 0\%$ \\
Other (data loading, sync, logging) & 4.80  & 6.61  & $+1.81$ & $+2.0\%$ \\
\midrule
\textbf{Total per step}& \textbf{92.40} & \textbf{130.83} & $\mathbf{+38.43}$ & $\mathbf{+41.6\%}$ \\
\bottomrule
\end{tabular}
\end{table}

\paragraph{Note on the per-step vs.\ end-to-end gap.} 
The per-step overhead ($+41.6\%$) measures only \texttt{update\_policy}'s post-warmup contribution; the end-to-end overhead ($+15.5\%$) is smaller because two amortization effects compound. 
(i) The $T_\text{warm}=50$ warmup steps carry no SD-Search overhead, so averaged over the full 200-step run the training-only overhead is $(24{,}244.5 - 18{,}480) / 18{,}480 = 31.2\%$, already below the per-step figure. 
(ii) Periodic validation, checkpoint saving, and setup, which together account for $\approx 5.2$h of AR's 10.3h end-to-end wall-clock and are identical between the two methods, inflate both totals equally and dilute the relative overhead from $31.2\%$ to $(11.9 - 10.3) / 10.3 = 15.5\%$.
Both effects shrink the gap as expected.

\paragraph{Comparison with process-supervision baselines.} The $15.5\%$ end-to-end overhead is contained entirely within the standard RL training loop: no external model inference, no annotation pipeline, no additional supervised stage. By contrast, the process-supervision baselines we compare against in Table~\ref{tab:main} carry substantial costs that do not appear in their reported training time:
\begin{itemize}
    \item \textbf{Thinker}~\citep{xu2025thinker} runs Qwen2.5-72B inference over the training set to generate sub-question decompositions, then performs a supervised fine-tuning stage on the resulting trajectories before RL. The 72B forward pass alone is $\sim 24\times$ the per-token cost of our 3B teacher pass on a parameter-count basis, applied to every training trajectory.
    \item \textbf{StepSearch}~\citep{wang2025stepsearch} relies on GPT-4o-generated sub-question annotations on the full $\sim 100\text{k}$-example NQ+HotpotQA training set, incurring both wall-clock and monetary costs that scale directly with dataset size.
\end{itemize}
SD-Search trades these multi-stage external pipelines for a single in-loop overhead, which we view as the more favorable point on the cost-quality frontier for practitioners without access to a 72B teacher or commercial API budget.

\paragraph{What end-to-end wall-clock includes.} 
The end-to-end figure (10.3h vs.\ 11.9h) covers the entire training run, including periodic validation evaluation on the seven benchmarks, checkpoint saving, and initial setup, in addition to the per-step training time measured in Table~\ref{tab:cost_breakdown}. 
The validation cost is identical for both methods (no SD-Search forward at evaluation time), so it inflates the absolute wall-clock equally and yields an end-to-end overhead ($+15.5\%$) that is correctly smaller than the per-step training overhead ($+41.6\%$).

\clearpage
\section{Training Dynamics}
\label{app:dynamics}

\par
\begin{figure}[htbp]
\centering
\includegraphics[width=\linewidth]{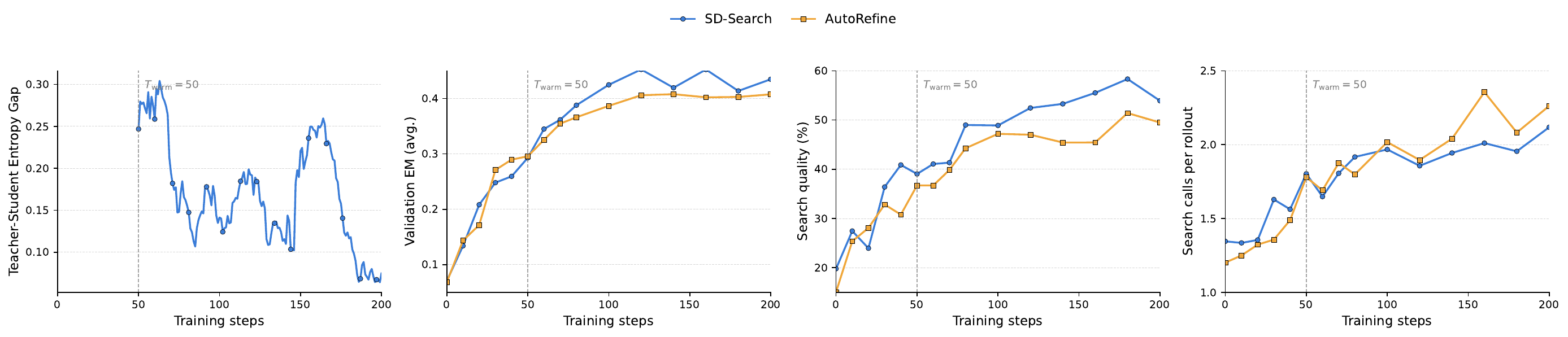}
\caption{Training dynamics of SD-Search versus AutoRefine on
Qwen2.5-3B-Base, averaged over the seven benchmarks. From left to right: the teacher-student entropy gap averaged over the search-query positions $\mathcal{Q}_\tau$ (only meaningful for SD-Search, since AutoRefine has no teacher view), validation Exact Match, search quality (the fraction of retrieved documents containing the gold answer), and search frequency (average number of search calls per rollout). The dashed line marks the warmup cutoff $\small T_{\mathrm{warm}}=50$, after which SD-Search's self-distillation loss is activated.}
\label{fig:training_dynamics}
\end{figure}

Figure~\ref{fig:training_dynamics} reports four training-time metrics on Qwen2.5-3B-Base, averaged across the seven benchmarks. 
The first metric, the teacher-student entropy gap on $\mathcal{Q}_\tau$ (the per-position difference $H(P^{\text{stu}}_p) - H(P^{\text{tch}}_p)$ averaged over search-query positions), is meaningful only for SD-Search, since AutoRefine has no teacher view. 
The remaining three, including validation Exact Match, search quality (the fraction of retrieved documents containing the gold answer), and search frequency (average number of search calls per rollout), are computed for both SD-Search and AutoRefine. 
We restrict the cross-method comparison to SD-Search and AutoRefine because both share the same pure-RL training schedule from the same base model; Thinker is trained with SFT on synthetic trajectories rather than RL from scratch, so its per-step training trajectory is not alignable to ours.

\par
Before the warmup cutoff $T_{\mathrm{warm}}=50$ SD-Search and AutoRefine are indistinguishable on EM, search quality, and search frequency, since SD-Search's distillation loss is not yet active. 
After step $50$ the curves diverge on EM and search quality but remain nearly overlapping on search frequency: SD-Search converges to roughly the same number of
search calls per rollout as AutoRefine, while achieving a higher fraction of successful retrievals and a higher final EM. 
The gain therefore comes from higher-quality queries at a comparable search rate, not from issuing more searches per question, ruling out the possibility that SD-Search's EM advantage is a byproduct of raw search volume or incidental training stabilisation.

\par
The entropy-gap panel is an interpretable proxy for the alignment that $\mathcal{L}_{\mathrm{SD}}$ in Eq.~\ref{eq:sd_loss} drives: as the JSD objective pulls the two distributions together their entropies must also converge. 
The gap is positive by construction, since the teacher's hindsight conditioning yields systematically sharper predictions than the student's prefix-only context. 
It hovers around $0.25$ pre-warmup, drops sharply to $\sim\!0.10$ within about $30$ post-warmup steps, and declines noisily toward $\sim\!0.065$ by step $200$, with a mid-training oscillation reflecting normal RL stochasticity. 
The gap does not reach zero: the teacher's hindsight block carries information the student cannot recover from its prefix, so a positive plateau is the expected steady state.

\clearpage
\section{Case Study}
\label{app:case}

Table~\ref{tab:case_study} shows how SD-Search, AutoRefine, and Thinker roll out on the same multi-hop question from 2Wiki. 
Incorrect answers are marked in red and correct answers in green; the underlined tokens in each search query highlight the entity or predicate the query is targeting. The example illustrates how the three methods differ in query focus during the first hop, which in this case is the hop where the chain either succeeds or fails.

\begin{table}[htbp]
\centering
\caption{Rollout comparison on a 2Wiki question requiring a bridge entity and a date lookup. }
\label{tab:case_study}
\small
\setlength{\tabcolsep}{4pt}
\begin{tabular}{p{0.13\linewidth} p{0.82\linewidth}}
\toprule
\multicolumn{2}{l}{\textbf{Question}: \quad What is the date of death of Alexander of Masovia's father?} \\
\multicolumn{2}{l}{\textbf{Gold Answer}: \quad 21 January 1426}. \\
\midrule
\multicolumn{2}{l}{\textit{\textbf{AutoRefine}}} \\
\midrule
\texttt{<think>} & I need to find the date of death of Alexander of Masovia's father. \\
\texttt{<search>} & \underline{Alexander of Masovia father date of death} \\
\texttt{<documents>} & [Doc 1: Alexander of Masovia] Alexander of Masovia (1400--2 June 1444) \ldots was the second son of Siemowit IV, Duke of Masovia \ldots \\
\texttt{<answer>} & \textcolor{red}{1400} \\
\midrule
\multicolumn{2}{l}{\textit{\textbf{Thinker}}} \\
\midrule
\texttt{<think>} & To answer this, I need to know who Alexander of Masovia's father is, then his death date. \\
\texttt{<search>} & \underline{Step1: Who is the father of Alexander of Masovia?} \\
                 & \underline{Action1: Retrieval(s=s1:person[`Alexander of Masovia'], p=p1:father, o=o1:person)} \\
\texttt{<documents>} & [Doc 1: Alexander of Masovia] \ldots was the second son of Siemowit IV, Duke of Masovia \ldots \\
\texttt{<search>} & \underline{Step2: Biography of \#1} \\
                 & \underline{Action2: Retrieval(s=o1, p=p2:biography, o=o2:biography)} \\
\texttt{<documents>} & [Doc 1: Siemowit IV] Siemowit IV (c.~1352--1426) was Duke of Masovia from 1381 until his death \ldots \\
\texttt{<answer>} & \textcolor{red}{1426} \\
\midrule
\multicolumn{2}{l}{\textit{\textbf{SD-Search}}} \\
\midrule
\texttt{<think>} & First identify the father, then look up his death date. \\
\texttt{<search>} & \underline{Alexander of Masovia father} \\
\texttt{<documents>} & [Doc 1: Alexander of Masovia] \ldots was the second son of Siemowit IV, Duke of Masovia \ldots \\
\texttt{<search>} & \underline{Siemowit IV Duke of Masovia date of death} \\
\texttt{<documents>} & [Doc 1: Siemowit IV] \ldots (c.~1352 -- 21 January 1426) \ldots \\
\texttt{<answer>} & \textcolor{green}{21 January 1426} \\
\bottomrule
\end{tabular}
\end{table}

The three methods diverge in how they decompose the question into search queries. 
AutoRefine collapses the two-hop question into a single query containing both the bridge entity placeholder and the target predicate; the retriever returns a document about Alexander himself, and the model extracts Alexander's own dates rather than his father's. 
Thinker correctly identifies the need for two hops, but its second query is a generic biography request that returns only a coarse year, missing the precise date. 
SD-Search issues two focused queries, each targeting a single fact (first the father, then the father's date of death), and recovers the full answer. 
The pattern is consistent with the training-dynamics finding that SD-Search's gain comes from query quality rather than query volume: both Thinker and SD-Search issue two searches, but SD-Search's queries are more surgically aimed at the fact being retrieved.

\clearpage
\section{Per-Token Distributional Effect of Hindsight Conditioning}
\label{app:token_trace}

\begin{figure}[!htbp]
\centering
\includegraphics[width=\linewidth]{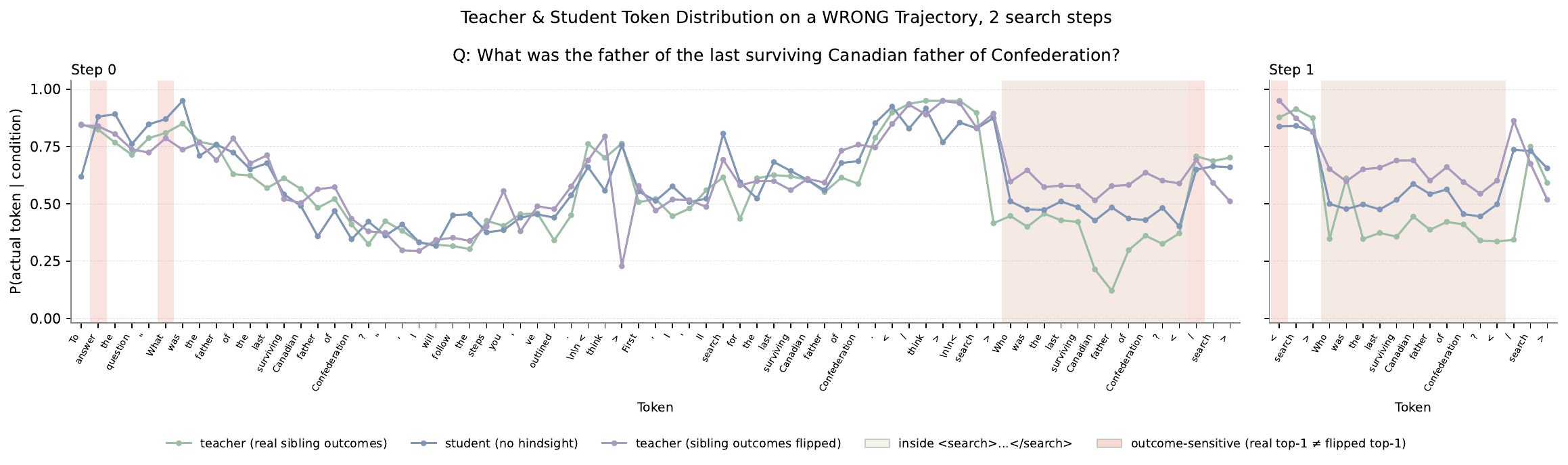}
\caption{Per-token trace of $P(\text{actual token}\mid\text{condition})$ across the action span of a failed Bamboogle rollout, in which the focal trajectory issues two near-identical queries, a degenerate failure mode that hindsight conditioning is designed to penalize.
\textbf{Student} (blue) sees no hindsight; \textbf{teacher with real outcomes} (green) is the SD-Search teacher view; \textbf{teacher with flipped outcomes} (purple) is a control in which every sibling rollout's outcome label is inverted. 
Cream bands mark search-query tokens (the alignment scope $\mathcal{Q}_\tau$); pink bands mark positions where the real and flipped teachers disagree on top-1.}
\label{fig:hindsight_trace}
\end{figure}

The mechanism in \S\ref{sec:hindsight} attributes SD-Search's step-level signal to the teacher redistributing probability mass at search-query positions when conditioned on hindsight. We probe this attribution directly by tracing $P(a_p \mid \text{condition})$ across a single rollout's action span, where $a_p$ is the token actually emitted at position $p$. Figure~\ref{fig:hindsight_trace} reports the trace under three conditions: \emph{student} (no hindsight, the standard inference-time context), \emph{teacher with real outcomes} (the full SD-Search teacher view), and \emph{teacher with flipped outcomes} (a control in which every sibling rollout's binary outcome label is inverted, so the hindsight block claims that focal-style queries are \textsc{Correct}). The example is a Bamboogle question on which the focal rollout fails; we plot the first two search steps.

Outside the search-query region the three conditions track each other with small natural fluctuation: hindsight has nothing to act on at the prompt prefix, the \texttt{<think>} span, or structural tags, since the prediction at these positions is already determined by the local reasoning context. Inside the search-query region (cream-shaded), the three traces separate consistently. The teacher under real outcomes assigns the lowest probability to the focal's actually-emitted query tokens, the teacher under flipped outcomes assigns the highest, and the student sits between them. The real-vs-flipped gap therefore localizes precisely on the tokens our objective targets ($\mathcal{Q}_\tau$), and leaves non-query positions almost unaffected. This is the signature SD-Search relies on: the JSD term in Eq.~\ref{eq:sd_loss} pulls the student toward the real-teacher distribution exactly where the hindsight signal is informative, while individual-token outliers (e.g., a single position at which one condition spikes due to a rare alternative being preferred) average out across the span.

\clearpage
\section{Scaling Across Model Sizes}
\label{app:scaling}
 
\begin{figure}[!htbp]
\centering
\includegraphics[width=0.5\linewidth]{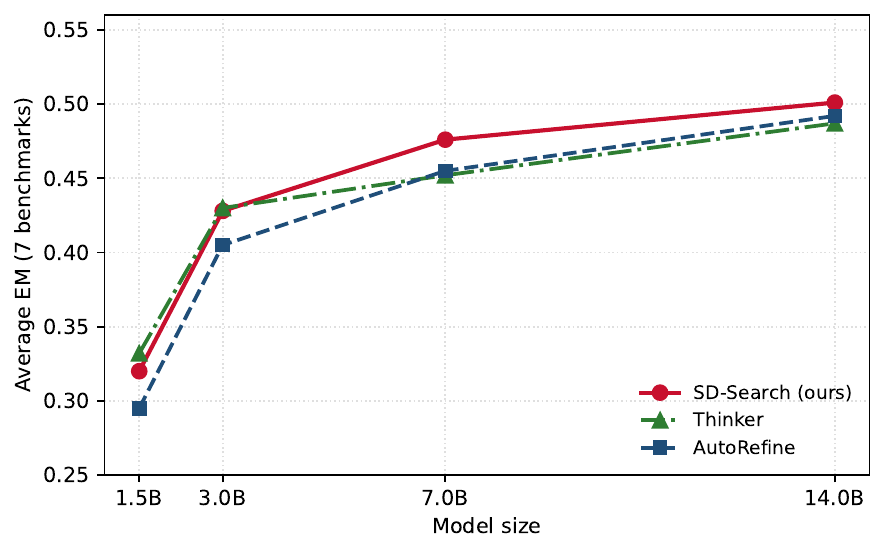}
\caption{
Scaling on Qwen2.5 at 1.5B, 3B, 7B, 14B (average EM across the seven benchmarks at training step 200). Thinker leads at 1.5B and ties SD-Search at 3B, but its gain over AutoRefine drops from $+3.7$ to $-0.5$ as scale grows from 1.5B to 14B. SD-Search's gain over AutoRefine stays positive at all four scales, narrowing only at 14B where higher base success rates reduce the within-group outcome contrast (see \S\ref{sec:limitations}).
}

\label{fig:scaling}
\end{figure}

We train SD-Search, Thinker, and AutoRefine on Qwen2.5 at four scales
(1.5B, 3B, 7B, 14B) under identical settings and report average EM across the seven benchmarks at step 200 in Figure~\ref{fig:scaling}.
The 3B and 7B points reproduce Tables~\ref{tab:main}
and~\ref{tab:main_7b}; the 1.5B and 14B points are trained with the same recipe.
 
The relative position of the three methods changes systematically with scale.
At 1.5B, both supervision-augmented methods clearly improve over AutoRefine ($+3.7$ for Thinker and $+2.5$ for SD-Search), with Thinker slightly ahead of SD-Search ($-1.2$); a 1.5B student appears to benefit more from a stronger external teacher than from its own hindsight view, since its in-context reading capacity is still limited.
At 3B the two methods become essentially indistinguishable
(SD-Search $0.428$ vs.\ Thinker $0.430$), with both improving over AutoRefine by $\approx{+}2.4$.
From 7B onward the ordering inverts: Thinker's gain over AutoRefine turns mildly negative ($-0.3$ at 7B and $-0.5$ at 14B), consistent with the saturation argument in \S\ref{sec:main_results}: as the student grows toward the 72B teacher, distillation from a fixed external reference delivers diminishing returns and at some point becomes a soft ceiling on the student. 
In contrast, SD-Search stays above AutoRefine at every scale. Comparing the best-performing variant of each method against AutoRefine-Base, the gain is $+2.5/{+}2.3/{+}2.1/{+}0.9$ at 1.5B/3B/7B/14B; the apples-to-apples Base-to-Base comparison is $+2.5/{+}2.3/{+}1.6/{+}0.6$ across the same scales (the 7B-Base gap of $+1.6$ widens to $+2.1$ once the SD-Search-Instruct variant is included, see Table~\ref{tab:main_7b}). Under both reportings the margin narrows at 14B but never crosses zero, because the teacher view is produced by the student itself and grows with the student's own in-context reading capacity rather than against a fixed reference.
 
\clearpage
\section{Teacher Input Schematic}
\label{app:teacher_input}

\S~\ref{sec:hindsight} formalizes the teacher view as $c^{\text{tch}}(\tau_i; \mathcal{G}) = [h(\mathcal{G}); \tau_i]$, which makes the conditioning structure clear but abstracts away what the teacher's input literally looks like as a token sequence. 
We provide a concrete instance in Figure~\ref{fig:teacher_input}: a single teacher input rendered as the model receives it, on the same 2Wiki question used in Appendix~\ref{app:case} under our default $G{=}5$ configuration. 
The figure shows the four blocks the teacher sees in causal order: the original \emph{prompt prefix} (identical to the student input up to the supervised step), a \emph{hindsight block} that prepends the $G{-}1$ sibling rollouts' search-only skeletons together with their outcome labels (after a string-level deduplication on the (queries, outcome) tuple), the focal trajectory's \emph{own future search spans} under future masking $\mathcal{M}$, and the focal trajectory's own \emph{outcome label}, after which the model predicts the next token of the supervised search query.

\begin{figure}[!htbp]
\centering

\begin{tabular}{|p{0.96\linewidth}|}
\hline

\rowcolor{gray!10}
\textbf{(1) Prompt prefix} \hfill {\small\textit{identical for student and teacher}} \\
\hline
\rowcolor{gray!10}
\small\ttfamily
You are a helpful assistant\ldots\ Question: What is the date of death of
Alexander of Masovia's father?\\
\rowcolor{gray!10}
\small\ttfamily
\textcolor{black!60}{<think>} First identify the father, then look up his death date. \textcolor{black!60}{</think>}\\
\rowcolor{gray!10}
\small\ttfamily
\textcolor{black!60}{<search>} \textcolor{blue!70!black}{Alexander of Masovia father} \textcolor{black!60}{</search>}\\
\rowcolor{gray!10}
\small\ttfamily
\textcolor{black!60}{<documents>} [Doc 1: Alexander of Masovia] \ldots second son of Siemowit IV, Duke of Masovia \ldots \textcolor{black!60}{</documents>}\\
\hline
\rowcolor{yellow!18}
\textbf{(2) Hindsight block:} sibling rollouts after dedup \\
\hline
\rowcolor{yellow!18}
\small\ttfamily
\textbackslash n[Trajectory Hindsight]:\textbackslash n\\
\rowcolor{yellow!18}
\small\ttfamily
[Sibling Rollout]: \textcolor{blue!70!black}{<search>Alexander of Masovia parents</search>}\ -\textgreater\ %
\textcolor{blue!70!black}{<search>Siemowit IV death</search>}\ %
\textcolor{igpositive}{[Outcome]: Correct}\\
\rowcolor{yellow!18}
\small\ttfamily
[Sibling Rollout]: \textcolor{blue!70!black}{<search>Alexander of Masovia father date of death</search>}\ %
\textcolor{red!75!black}{[Outcome]: Incorrect}\\
\rowcolor{yellow!18}
\small\ttfamily
[Sibling Rollout]: \textcolor{blue!70!black}{<search>father of Alexander of Masovia</search>}\ -\textgreater\ %
\textcolor{blue!70!black}{<search>Siemowit IV biography</search>}\ %
\textcolor{red!75!black}{[Outcome]: Incorrect}\\
\rowcolor{yellow!18}
\small\ttfamily
[Sibling Rollout]: \textcolor{blue!70!black}{<search>Alexander of Masovia</search>}\ -\textgreater\ %
\textcolor{blue!70!black}{<search>Siemowit IV Duke of Masovia</search>}\ -\textgreater\ %
\textcolor{blue!70!black}{<search>Siemowit IV January 1426</search>}\ %
\textcolor{igpositive}{[Outcome]: Correct}\\
\hline

\rowcolor{yellow!18}
\textbf{(3) Focal future} (own search spans only, future-masked $\mathcal{M}$) \\
\hline
\rowcolor{yellow!18}
\small\ttfamily
\textcolor{blue!70!black}{<search>Siemowit IV Duke of Masovia date of death</search>}\\
\hline

\rowcolor{yellow!18}
\textbf{(4) Focal outcome label} \\
\hline
\rowcolor{yellow!18}
\small\ttfamily
\textbackslash n\textcolor{igpositive}{[Outcome]: Correct}\textbackslash n\\
\hline

\rowcolor{cyan!10}
\textbf{(5) Supervised action span $\boldsymbol{\tau_{i,p:p+|a|}}$} \hfill {\small\textit{JSD applied at $\mathcal{Q}_\tau$ tokens}} \\
\hline
\rowcolor{cyan!10}
\small\ttfamily
\textcolor{black!60}{<search>} \textcolor{blue!70!black}{\underline{Siemowit IV Duke of Masovia date of death}} \textcolor{black!60}{</search>}\\
\hline
\end{tabular}
\caption{Teacher input layout for one supervised step. Blocks (1) and (5) are
the standard student input split around the action being supervised.
Blocks (2)--(4) are inserted between them and constitute the hindsight
context $h(\mathcal{G})$: the $G{-}1$ sibling rollouts in (2) (search
spans only, each tagged with its outcome label, with $G{=}5$ in our
default configuration), the focal trajectory's own future
search spans in (3) (other span types removed by future masking), and
the focal trajectory's own outcome label in (4). \textcolor{blue!70!black}{Blue}
tokens are search-query content, the positions $\mathcal{Q}_\tau$ at
which the JSD objective in Eq.~\ref{eq:sd_loss} is applied; outcome
labels are colored \textcolor{igpositive}{green} for \textsc{Correct}
and \textcolor{red!75!black}{red} for \textsc{Incorrect}.
At inference time the yellow-shaded blocks (2)--(4) are absent and the
teacher view collapses back to the student view.}
\label{fig:teacher_input}
\end{figure}

\par
Three details of this layout are worth noting. 
First, this is the concrete token-level realization of the abstract context $c^{\text{tch}}(\tau_i;\mathcal{G}) = [h(\mathcal{G});\tau_i]$ in Eq.~\ref{eq:sd_teacher_ctx}: the hindsight block is placed immediately before the supervised action positions while keeping the question prefix unchanged relative to the student input, so the diff between teacher and student contexts at each supervised position is exactly the highlighted hindsight block, and every supervised position has causal access to all of $h(\mathcal{G})$.
Second, the focal trajectory's own future search spans (block 3) are included in the hindsight, separated from the sibling rollouts; this is what makes the leave-one-out variant in \S\ref{sec:ablation_teacher} a meaningful probe, since it removes exactly this block while preserving block 2.
Third, future masking $\mathcal{M}$ is applied uniformly to both the sibling skeletons (block 2) and the focal future (block 3), so that \textsc{think} reasoning, retrieved \textsc{documents}, and final \textsc{answer} spans are all stripped out before reaching the teacher; without this masking the teacher would see passages whose content makes the gold answer extractable directly, collapsing the search-position distributions onto retrieval-skipping continuations as quantified in \S\ref{sec:ablation_teacher}.


\clearpage

\end{document}